\documentclass[conference]{IEEEtran}
\IEEEoverridecommandlockouts
\usepackage{latexsym}
\usepackage{booktabs}
\usepackage{enumitem}
\usepackage{multicol}
\usepackage{multirow}
\usepackage{algorithm}
\usepackage{paralist}
\usepackage{cite}
\usepackage{amsmath,amssymb,amsfonts}
\usepackage{algorithmic}
\usepackage{graphicx}
\usepackage{textcomp}
\usepackage{xcolor}
\def\BibTeX{{\rm B\kern-.05em{\sc i\kern-.025em b}\kern-.08em
    T\kern-.1667em\lower.7ex\hbox{E}\kern-.125emX}}
\begin{document}

\title{Transferable Physical-World Adversarial Patches Against Object Detection in Autonomous Driving}

\author{
\IEEEauthorblockN{Zihui Zhu}
\IEEEauthorblockA{zhuzihui@hust.edu.cn\\
Huazhong University of Science and\\ Technology\\
China}
\and
\IEEEauthorblockN{Ziqi Zhou}
\IEEEauthorblockA{zhouziqi@hust.edu.cn\\
Huazhong University of Science and\\ Technology\\
China}
\and
\IEEEauthorblockN{Yichen Wang}
\IEEEauthorblockA{wangyichen@hust.edu.cn\\
Huazhong University of Science and\\ Technology\\
China}
\and
\IEEEauthorblockN{Lulu Xue}
\IEEEauthorblockA{lluxue@hust.edu.cn\\
Huazhong University of Science and\\ Technology\\
China}
\and
\IEEEauthorblockN{Minghui Li}
\IEEEauthorblockA{minghuili@hust.edu.cn\\
Huazhong University of Science and\\ Technology\\
China}
\and
\IEEEauthorblockN{Shengshan Hu}
\IEEEauthorblockA{hushengshan@hust.edu.cn\\
Huazhong University of Science and\\ Technology\\
China}
}

\maketitle

\begin{abstract}
% Deep learning drives substantial progress in safety-critical domains such as autonomous driving, where object detection models serve as core components in vehicle perception systems.
% Existing physical-world adversarial patch attacks targeting object detectors are typically optimized for individual models, limiting their transferability to unknown detectors.
% In this paper, we introduce AdvAD, a transfer-oriented physical-world adversarial attack that disrupts the object detection in autonomous driving. AdvAD adopts a multi-model ensemble strategy and introduces adaptive model weighting, which dynamically adjusts the contribution of each model in a bi-directional query loop based on its gradient utility in each iteration.
% In addition, we incorporate data augmentation and geometric transformation techniques to enhance patch robustness under varied physical-world conditions.
% Extensive experiments in both digital and physical settings show that AdvAD achieves superior attack performance, transferability, and robustness compared to state-of-the-art methods.
% The code is available at \url{https://github.com/Zhou-Zi7/AdvAD}

Deep learning drives major advances in \textit{autonomous driving} (AD), where object detectors are central to perception.
However, adversarial attacks pose significant threats to the reliability and safety of these systems, with physical adversarial patches representing a particularly potent form of attack.
Physical adversarial patch attacks pose severe risks but are usually crafted for a single model, yielding poor transferability to unseen detectors.
We propose AdvAD, a transfer-based physical attack against object detection in autonomous driving.
Instead of targeting a specific detector, AdvAD optimizes adversarial patches over multiple detection models in a unified framework, encouraging the learned perturbations to capture shared vulnerabilities across architectures. The optimization process adaptively balances model contributions and enforces robustness to physical variations.
% , enabling the generated patches to remain effective under diverse viewpoints and environmental conditions.
% AdvAD ensembles multiple models with adaptive weighting, dynamically updating each model’s influence in a bi-directional query loop based on gradient utility.
It further employs data augmentation and geometric transformations to maintain patch effectiveness under diverse physical conditions.
Experiments in both digital and real-world settings show that AdvAD consistently outperforms state-of-the-art (SOTA) attacks in performance and  transferability.

\end{abstract}

% \begin{keywords}
% Object Detection, Adversarial Patch, Physical-World Attack
% \end{keywords}


\section{Introduction}
\label{sec:1}
\begin{figure}[t]
    \centering
    \includegraphics[scale=0.26]{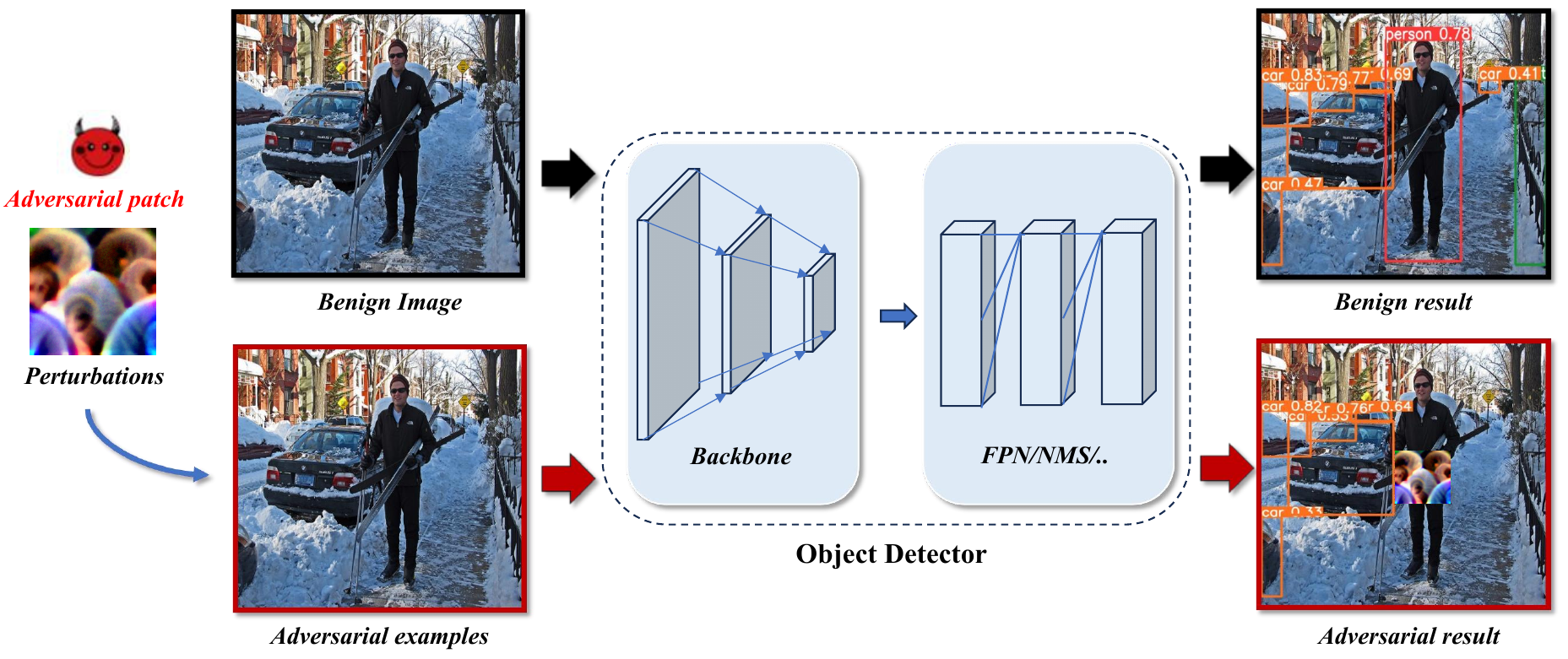}
    \caption{An illustration of attacking object detection models in autonomous driving systems using adversarial patches. The adversarial patch is strategically placed in the environment, leading the detector to miss safety-critical objects and potentially compromising perception reliability.
    }
    \label{fig1:demo}
      \vspace{-0.4cm}
\end{figure}

With the rapid development of deep learning~\cite{xue2025towards,pan2026ufvideo,wu2025tattoo}, object detection models have been widely deployed in critical real-world applications \cite{Sallab2017DeepRL, martinez2017object}. 
% such as autonomous driving \cite{Sallab2017DeepRL}, intelligent surveillance  \cite{valera2005intelligent}, and robotics \cite{martinez2017object}. 
% Ensuring their reliability is therefore of great importance.
% However, 
% Recent studies have demonstrated that object detectors are vulnerable to adversarial attacks~\cite{huang2017adversarial}, where carefully designed perturbations mislead predictions and pose severe security risks.
Recent studies show that these detectors are exposed to numerous security threats~\cite{zhou2025darkhash, badhash,wan2025mars,zhang2024detector,zhang2025test,wang2024trojanrobot,li2025detecting,wang2024eclipse,yu2025spa,wang2024unlearnable}, particularly exhibiting high vulnerability to adversarial attacks \cite{zhou2024securely,advclip,zhou2023downstream,zhou2025numbod,zhou2024darksam}, where carefully designed perturbations mislead predictions.
Among various approaches, as illustrated in Fig.~\ref{fig1:demo}, patch-based adversarial attacks \cite{Wang2019advPatternPA,Lin2024OutofBoundingBoxTA} have attracted particular attention due to their practicality.
% Patch-based attacks have attracted increasing attention due to their practicality and robustness, but recent studies \cite{Lin2024OutofBoundingBoxTA, Huang2022TSEATS} reveal two critical limitations: limited transferability to black-box models with different architectures and significant performance degradation under environmental changes such as varying viewpoints, illumination, and sensor noise.

Despite their effectiveness, existing patch-based attacks face two major challenges:
1) Poor transferability. 
Their transferability to black-box detectors with different architectures remains limited. 
Existing ensemble-based methods \cite{Huang2022TSEATS,Cai2023EnsemblebasedBA} attempt to mitigate this by aggregating gradients from multiple models, but uniform weighting often leads to overfitting.
DOEPatch \cite{tan2024doepatch} improves transferability through dynamically adjusted weights in a min–max framework, yet its optimization is primarily energy-driven rather than detection-aware. Although the perturbations achieve broad spectrum coverage, they may fail to sufficiently exploit the intrinsic vulnerabilities of detection pipelines, limiting their effectiveness against real detectors.
2) Environmental sensitivity.
Physical-world evaluation has been restricted to limited scenarios, leaving robustness under diverse conditions largely unexplored \cite{liu2025hawk}. Adversarial perturbations deployed in real-world scenarios are easily affected by environmental factors such as viewing angles, lighting variations, and weather conditions, which can significantly reduce their effectiveness \cite{zheng2025blackboxbench}. This sensitivity arises because most perturbations are optimized under idealized digital conditions with fixed viewpoints and clean inputs. When deployed physically, the perturbation undergoes various transformations, which distort its adversarial features and weaken its ability to mislead detectors.
% Physical-world evaluation has been restricted to limited scenarios, leaving robustness under diverse conditions largely unexplored. Adversarial perturbations in the real world are easily affected by environmental factors such as viewing angles, lighting variations, weather conditions, and sensor noise, which can significantly reduce their effectiveness.}

%

\begin{figure*}[!h]
    \centering
    \includegraphics[scale=0.6]{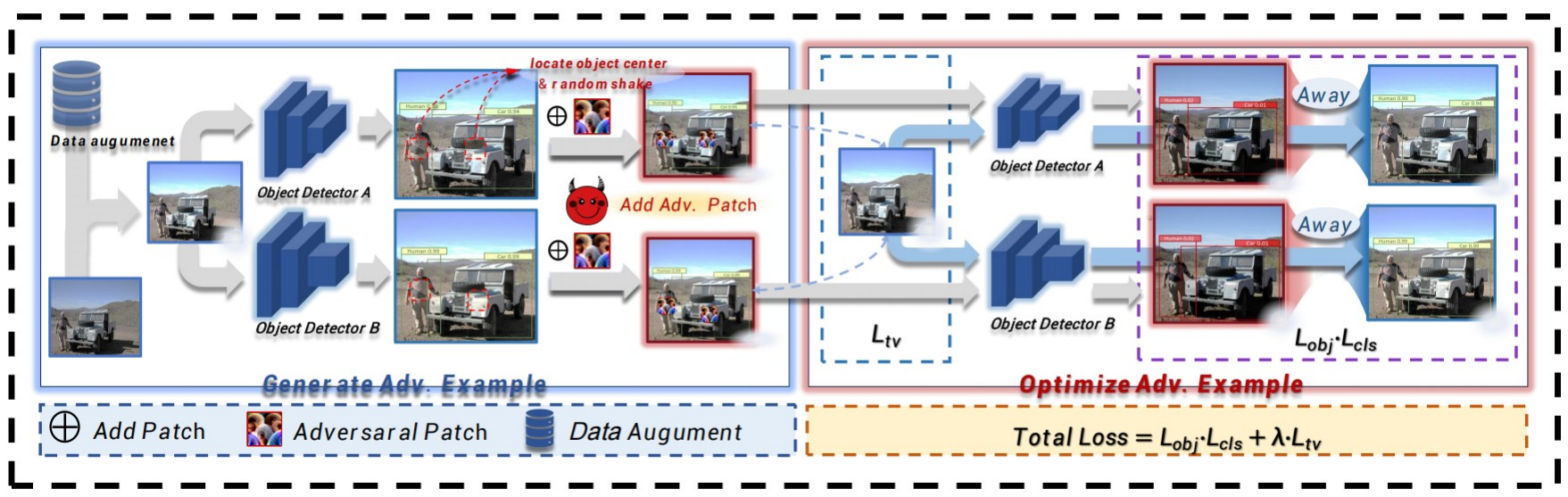}
    \caption{Overall framework of the proposed AdvAD method. Given an input image, tailored data augmentation is first applied to simulate diverse real-world physical conditions. Two pre-trained white-box object detectors are then employed to localize target objects in the augmented images, and the adversarial patch is placed at the center of each detected object with a fixed relative scale to the object size. This framework enables the generation of physical adversarial patches that remain effective under diverse real-world conditions.
    }
    \label{fig:pipeline}
    
\end{figure*}

In this paper, we propose AdvAD, a novel adaptive physical-world adversarial attack designed for autonomous driving systems. 
Unlike prior energy-based ensemble methods, AdvAD integrates both feature-level and output-level gradients from multiple surrogate detectors and introduces a detection-aware adaptive weighting strategy that adaptively balances contributions across models to mitigate overfitting. 
Moreover, we design a realistic deployment augmentation suite that incorporates random cropping, geometric transformations, illumination changes, and total variation regularization to simulate real-world distortions to sustain high attack success rates under diverse deployment conditions.

Our main contributions are summarized as follows:
\begin{inparaenum}[\hspace{0.5em}1)]
\item We propose AdvAD, an adversarial patch framework that simultaneously achieves high transferability and physical robustness for object detection tasks.
\item We introduce a detection-aware dynamic weighting strategy, which adaptively balances model contributions and mitigates overfitting in ensemble training.
\item We conduct extensive experiments on multiple benchmarks and detectors. Results demonstrate that {AdvAD} consistently outperforms existing SOTA methods in terms of attack effectiveness and transferability.
\end{inparaenum}

\section{Related Work}
\label{sec:relat}
\subsection{Object Detection}
% Object detection methods \cite{dalal2005histograms,lowe1999object,freund1997decision} typically extracted features that often lack generalization and robustness in complex real-world environments, rendering these methods insufficient for current detection requirements.
Early object detection methods \cite{dalal2005histograms,lowe1999object,freund1997decision} relied on hand-crafted features, which lack generalization and robustness in complex real-world environments, making them inadequate for modern detection tasks.
Recently, deep learning-based object detection algorithms have advanced rapidly, with two-stage (e.g., SPP-Net~\cite{he2015sppnet}, Faster R-CNN~\cite{ren2015fasterrcnn}, Cascade R-CNN~\cite{cai2018cascade}).
% Recently, deep learning-based object detection algorithms have advanced rapidly, with two-stage(e.g., SPP-Net~\cite{he2015sppnet}, Faster R-CNN~\cite{ren2015fasterrcnn}, Cascade R-CNN~\cite{cai2018cascade} ) and single-stage detectors (e.g., YOLO~\cite{khanam2024yolov11}, SSD~\cite{liu2016ssd}, DETR~\cite{carion2020detr}) becoming the standard. 
% Two-stage detectors, such as SPP-Net~\cite{he2015sppnet}, Faster R-CNN~\cite{ren2015fasterrcnn}, Mask R-CNN~\cite{he2017maskrcnn}, and Cascade R-CNN~\cite{cai2018cascade}, generate candidate regions through a region proposal network (RPN) and then perform classification and bounding box regression on these regions. Although this approach achieves high accuracy, it suffers from relatively lower efficiency. 
% Two-stage detectors, generate candidate regions through a region proposal network (RPN) and then perform classification and bounding box regression on these regions. Although this approach achieves high accuracy, it suffers from relatively lower efficiency.
In contrast, single-stage detectors, including the YOLO~\cite{khanam2024yolov11}series, SSD~\cite{liu2016ssd}, RetinaNet~\cite{lin2017retinanet}, and DETR~\cite{carion2020detr}, eliminate the region proposal stage to improve detection efficiency, performing category classification and bounding box regression directly on the input image.
Among these, the YOLO series has seen continuous improvements in both detection accuracy and inference speed. 
% Notably, YOLOv5 has gained widespread adoption in real-world applications due to its remarkable balance between performance and efficiency. 
% The latest versions, YOLOv10 and YOLOv11, offer enhanced accuracy and even faster inference speeds.

\subsection{Adversarial Attacks on Object Detection}
Adversarial attacks mislead detectors by injecting carefully designed perturbations. Existing approaches can be broadly categorized into perturbation-based \cite{zhou2025sam2,wang2025advedm,wang2025breaking,song2025segment,li2024transferable,song2025seg,song2026erosion,Carlini2016TowardsET} and patch-based \cite{Brown2017AdversarialP,Lee2021AntiAdversariallyMA,11209924,hu2025two} methods.
While effective in digital settings, their success degrades in the physical world due to viewpoint shifts, lighting variations, and printing artifacts.
To bridge this gap, recent work explores robust physical attacks using printed stickers \cite{etim2025adversarial}, adversarial paint \cite{Suryanto2023ACTIVETH}, or optical projection \cite{Huang2022TSEATS},or 3D-based adversarial attacks \cite{wang2025unified}. 
Recent work \cite{tiliwalidi2025adversarial} has also investigated translucent adversarial patches to improve the stealth of physical attacks on object detectors, demonstrating their effectiveness in both digital and physical settings.
Despite these advances, such methods remain sensitive to environmental conditions and exhibit poor cross-model transferability, limiting their practicality.

\section{Methodology}\label{sec:Methodology}
% In this section, we introduce a new approach to adversarial patch generation that has been made AdvAD.

\begin{figure*}[!t]
    \centering
    \includegraphics[scale=0.5]{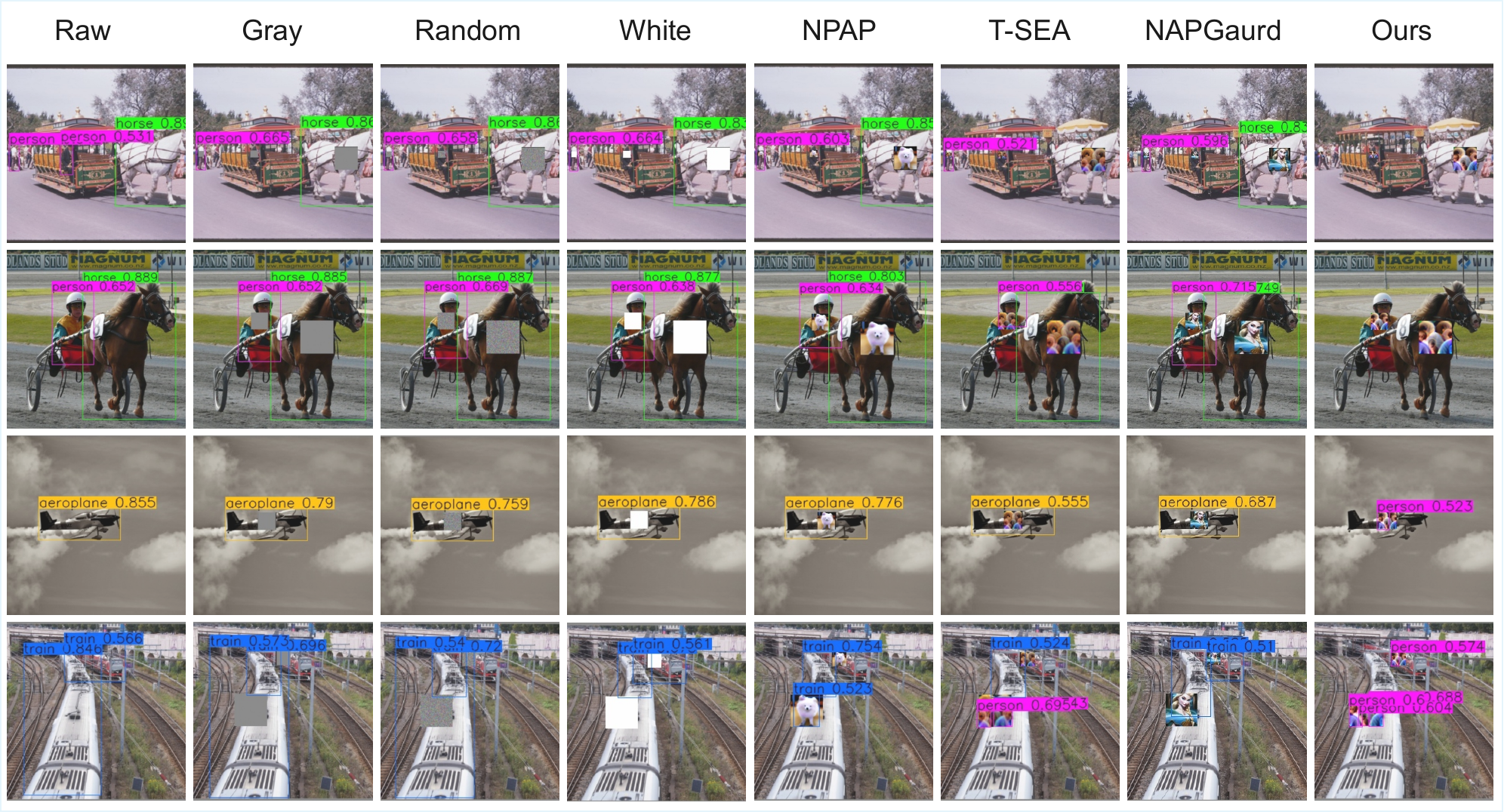}
    \caption{Visualization results in the digital world. The adversarial patch is aligned with the target object region during attack generation. Compared with baseline methods, AdvAD results in more missed detections and reduced detection confidence. These visual results demonstrate the effectiveness of AdvAD in degrading object detector performance in the digital domain.
    }
    \label{fig3:Attack Performance}
\end{figure*}

\subsection{Problem Formulation}
% In this work, we focus on physical-world adversarial patch attacks, where an attacker places adversarial patches in a real-world environment to spoof a object detection model deployed in an autonomous driving system. 
We formalize the task of generating physical adversarial patches for object detection. 
Let $\mathcal{X}$ be the space of raw images and $\mathcal{Y}$ the space of detection outputs. 
Object detector can be written as
$f:\mathcal{X}\to\mathcal{Y}, \quad f = h \circ g \circ S ,$
where $S$ is the preprocessing stage (resize, normalize, pad), 
$g$ is the feature extractor, 
and $h$ is the detection head producing a set of predictions
$y=\{(b_i,c_i,s_i)\}_{i=1}^N$, 
with $b_i$ is bounding box, $c_i$ is class label, and $s_i$ the confidence score.

\noindent\textbf{Threat model.} The attacker aims to produce a physical patch 
$p$ that, when physically realized and placed in the scene, causes the target detector 
$f$ to make specified errors. Physical adversarial patch $p\in\mathbb{R}^{w_p\times h_p\times 3}$, when place in the scene and observed by a camera, generates an adversarial input via the operator
$ x_{\text{adv}} = \mathcal{A}(x,p,\tau), $
where $\tau \in \mathcal{T}$ encodes physical transformations such as 
scale, rotation, viewpoint, and illumination. 
The detector output on this adversarial image is
\begin{equation}
y_{\text{adv}} = f(x_{\text{adv}}) = h\bigl(g(S(\mathcal{A}(x,p,\tau)))\bigr).
\end{equation}

\subsection{Overall Framework}
Our proposed method, the AdvAD pipeline, as illustrated in 
Fig.\ref{fig:pipeline}, aims to generate an adversarial patch that can maintain its attack effectiveness against the object detector in real-world autonomous driving scenarios. 
Given an input image $x$, we first apply tailored data augmentation to mimic dynamic real-world conditions. 
Two pre-trained white-box detectors are then used to localize objects in the augmented images, and the adversarial patch is placed at the center of each detected object. 
Each detector contributes an individual loss, and adaptive optimization is performed by dynamically weighting these losses.
The patch is optimized by minimizing the following total loss:
\begin{equation}
    \mathcal{L}_{\text{total}} = \mathcal{L}_{\text{obj}} \cdot \mathcal{L}_{\text{cls}} + \lambda_{\text{tv}} \cdot \mathcal{L}_{\text{tv}} 
\end{equation}
%Where \( \mathcal{L}_{\text{obj}} \) denotes the target loss to measure the effect of the patch on the detector and \( \mathcal{L}_{\text{cls}} \) denotes the categorical loss.
% $\mathcal{L}_{\text{cls}}$ is the cross-entropy loss to enforce misclassification:
where \( \mathcal{L}_{\text{obj}} = \frac{1}{N} \sum_{i=1}^{N} C_i \) is the average object confidence and 
\(
    \mathcal{L}_{\text{cls}}
= - \sum_{c=1}^{C} y_{c} \log \left( \frac{e^{f_{c}}}{\sum_{j=1}^{C} e^{f_{j}}} \right)
\)
% is the cross-entropy loss encouraging misclassification. 
is the cross-entropy loss over $C$ classes, with $y_c$ the target label and $f_c$ the predicted detector score,
The total variation loss,
\(
\mathcal{L}_{\text{tv}} = \sum_{i,j} \left( \left| P_{i,j+1} - P_{i,j} \right| + \left| P_{i+1,j} - P_{i,j} \right| \right)
\), with a lower bound $\tau=0.1$ to avoid vanishing. $\lambda_{\text{tv}}$ controls its weight.

Finally, we apply patch cutout during training to mitigate overfitting to specific image regions.
% 
% are applied to the patch to prevent overfitting when the adversarial patch is applied to each object of the target class.

\subsection{Ensemble-based Adversarial Attack}

% \subsubsection{Multi-model Integration Attack}
\noindent\textbf{Multi-model Integration Attack.}
% Due to the limited information obtained from a single model, the adversarial examples have limited transferability between different black-box models, and the effect of the integrated attack can effectively improve the generalization ability of the adversarial samples compared to a single model that combines information from multiple models. 
% We therefore employ a multi-model integration attack to capture more generalized perturbation information to improve patch transfer. 
% In the integrated attack we use a collection of \( M \) white-box agent models  \( \mathcal{G} = \{ g_1, g_2, \dots, g_M \} \) to jointly generate an adversarial patch \( \hat{p} \), which is then used to attack the target black-box model \( g_{\text{victim}} \).
Adversarial examples generated from a single model usually exhibit poor transferability across black-box detectors. 
We therefore optimize a single patch across an ensemble of $M$ white-box agent models $\mathcal{G}=\{g_1,\dots,g_M\}$:
\begin{equation}
\hat{p} = \arg\min_{p\sim\mathcal{P}} \sum_{m=1}^{M} \alpha_m , \mathcal{L}_m(\tilde{x}_i, p),
\end{equation}
where $\alpha_m$ is the weight of model $g_m$ and $\tilde{x}_i $ is the augmented input image, and $\mathcal{L}_m(\tilde{x}_i, p)$ is the attack loss computed on model $g_m$ for patch $p$
Training aggregates losses from the ensemble to produce a patch with improved generalization.
% where \( \alpha_m \) is the weight of model \( g_m \), and \( \mathcal{L}(\tilde{x}_i, g_m, p) \) denotes its loss. Adversarial patches are trained sequentially on \( M \) detectors, and their losses are aggregated for optimization.
% where \( \alpha_m \) denotes the weight parameter of different white-box agent models, and \( \mathcal{L}(\tilde{x}_i, g_m, p) \) denotes the corresponding loss function of the white-box agent model. We train a generalized adversarial patch by executing attacks on \( M \) detectors sequentially and computing the loss of each detector.

% \subsubsection{Dynamic loss-weighting strategy}
\noindent\textbf{Dynamic Loss-weighting Strategy.}
% Previous integration approaches are set \( \alpha_m = \frac{1}{m}\) to generate adversarial samples by weighted averaging the gradient information of each model.
We adaptively tune $\alpha_m$ during training to avoid overly focus models with large losses.
During patch optimization, the ensemble weights $\alpha_m$ are adaptively adjusted to balance the influence of each surrogate model, while keeping the model parameters fixed.
Weights are initialized uniformly: \( \alpha_0^{(m)} = \frac{1}{M}, \quad \forall i \in \{1, 2, \dots, M\} \) 
and at each iteration, the weight $\alpha_m$ is adjusted according to the gradient of the total loss with respect to 
$\alpha_m$:
\begin{equation}
\alpha_m^{(t+1)} = \alpha_m^{(t)} - \eta \frac{\partial \mathcal{L}_{\text{total}}}{\partial \alpha_m^{(t)}},
\end{equation}
where $\eta$ is a small step size. After each update, the weights are normalized to ensure stability.

\subsection{Patch Cropping}
We apply random patch cropping during training to improve patch generalization and reduce reliance on specific regions. 
Instead of always using the full patch, we randomly crop a small rectangular region within the patch before applying it to the image. 
Given a normalized image \(x\) and patch \(P\), cropping is applied with probability \(p_{\text{crop}}\). 
% When triggered, we uniformly sample a square mask of side length \(s\) inside \(P\) and replace its pixels with a constant value \(v\) (e.g., 0 or the patch mean). 
% The masked patch is then applied to \(x\). 
A square region of side length  $s$ in $P$ is replaced with a constant value $v$ (e.g., 0 or the patch mean) before applying the patch to $x$.
This approach encourages the patch to distribute its adversarial effect across a wider area, rather than relying heavily on specific pixels.
\section{Experiments}
\subsection{Experimental Setup}

\begin{table*}[!t]   % twocolumn \begin{table*}
		\fontsize{8}{12}\selectfont   
		\centering
		\caption{Comparison of mAP (\%) for different attack methods on MS-COCO and PASCAL VOC.}
		\label{t1}
			\begin{tabular}{ccccccccc}
                \toprule[1pt]
                 \multirow{2}{*}{Methods}
                    & \multicolumn{4}{c}{MS-COCO} 
                    & \multicolumn{4}{c}{PASCAL VOC}
                  \\
                \cmidrule(lr){2-5} \cmidrule(lr){6-9}
                & YOLOv5  & SSD & Faster R-CNN &Avg.
                & YOLOv5  & SSD & Faster R-CNN &Avg.
                 \\
    \midrule[1pt]
    Gray          & 58.89 & 56.89 & 46.95 &54.24 & 48.03 & 49.30 & 35.85  &44.39 \\
    Random Noise  & 60.10 & 56.19 & 52.53 &56.27 & 47.39 & 47.16 & 41.21 & 45.25 \\
    White         & 53.75 & 52.07 & 45.61 &50.48 & 40.46 & 43.84 & 32.61 & 38.97 \\
    \midrule[1pt]
    AdvPatch ~\cite{thys2019fooling}      & 25.24 & 37.47 & 33.57 & 32.09 & \textbf{18.55} & 28.84 & 26.07 & 24.49\\
    NPAP ~\cite{hu2021naturalistic}         & 42.65 & 45.71 & 39.82 & 42.73 & 28.71 & 33.29 & 29.13 & 30.38 \\
    T-SEA ~\cite{Huang2022TSEATS}         & 28.22 & 40.06 & 34.41 & 34.23 & 21.93 & 32.78 & 26.92  & 27.21\\
%    AdvAD (Ours)   & \textbf{24.49} & \textbf{33.38} & \textbf{29.97} &\textbf{29.28}  & 19.15 & \textbf{26.31} & \textbf{22.69} & \textbf{22.72}\\
    NAPGuard ~\cite{wu2024napguard}       & 44.91  & 48.08  & 39.91  & 44.30  & 32.70  & 34.64  & 31.16   & 32.83 \\
    AdvAD (Ours)   & \textbf{23.61} & \textbf{32.08} & \textbf{29.27} &\textbf{28.32}  & 19.15 & \textbf{26.31} & \textbf{22.69} & \textbf{22.72}\\
    \bottomrule[1pt]
                \end{tabular}
        %        }
		%}
	\end{table*}   % twocolumn \end{table*}

\noindent\textbf{Datasets and models.} 
We train adversarial patches on the INRIA dataset~\cite{dalal2005histograms}. 
Evaluation is conducted on MS-COCO 2017~\cite{Lin2014MicrosoftCC} and PASCAL VOC~\cite{Everingham2010ThePV}, using both single- and two-stage detectors (YOLOv5, SSD, Faster R-CNN).

\noindent\textbf{Parameter setting.} 
For each image during training patches, we resize them to 416 × 416, and adversarial patches are set to 300 × 300.
% and then set the size of the generated adversarial patches to 300 × 300.
Data augmentation includes horizontal flipping, small-angle rotation, slight color perturbation, and random scale cropping.
% For the input images, we adopt a restricted set of column data enhancement strategies, including horizontal flipping, random small-angle rotation, slight color perturbation, and random scale cropping.
%, to achieve diversity within a limited range of transformations. 
% We set the hyperparameter  $\lambda_{\text{tv}}$ to 2.5, while the training epoch number is set to 1000 with a batch size of 16. The attack process is optimized by a maximum perturbation range of 255 ~\cite{Carlini2016TowardsET}. 
We set $\lambda_{\text{tv}}=2.5$, train for $1000$ epochs with batch size $16$, and clip pixel values to $255$ ~\cite{Carlini2016TowardsET}.
Optimization uses Adam with learning rate $0.03$; ALRS is applied for dynamic step-size adjustment.
% We set the hyperparameter $\lambda_{\text{tv}}$ to 2.5, train for 1000 epochs with a batch size of 16, and constrain perturbations within 255~\cite{Carlini2016TowardsET}.
% Optimization is performed using Adam~\cite{Kingma2014AdamAM} with a learning rate of 0.03. To improve stability, we adopt ALRS for dynamic step-size adjustment.
%We use Adam ~\cite{Kingma2014AdamAM} as the optimizer and set the learning rate to 0.03. To enhance the attack stability, we use ALRS in order to dynamically adjust the update step size.

\noindent\textbf{Evaluation metrics.} 
We report mean average precision (mAP), computed over all classes at an IoU threshold of 0.45~\cite{Huang2022TSEATS}. Attack strength is quantified by the mAP drop between clean and adversarial images.
% We use mean average precision (mAP) to measure attack effectiveness. Following~\cite{Huang2022TSEATS}, mAP is computed over all classes at an IoU threshold of 45\%. The performance drop between clean and adversarial images reflects the attack strength.
%We adopt mean average precision (mAP), to measure the attack performance of our designed patches. 
%We compare the map between the clean images and adversarial images with the patches. 
%Following ~\cite{Huang2022TSEATS}, we compute the mAP value over all detection classes, using an Intersection over Union (IoU) threshold of 45\%.
%A larger decrease in mAP indicates the more effective attack.

\subsection{Digital Domain Evaluation} \label{sec:attack_performance}
% To comprehensively evaluate the effectiveness of the AdvAD method, we investigate its attack performance against three state-of-the-art methods on two datasets and present the results in Table \ref{t1}. 
%For each attack, we generate adversarial examples by combining clean samples from the datasets with adversarial patches created by different methods and evaluate the effectiveness of AdvAD through detection by the detector.
We evaluate AdvAD against three SOTA baselines on two datasets in Table~\ref{t1}.
Adversarial patches are applied to clean images, and performance is measured by detection mAP.
AdvPatch and T-SEA are trained on YOLOv5s, while NPAP is trained on YOLOv4s.
% Adversarial examples are generated by applying patches to clean images, and performance is measured by detection mAP. 
% Where, AdvPatch and T-SEA are trained on YOLOv5s, while NPAP is trained on YOLOv4s. 
Results show that AdvAD achieves the largest mAP reduction, outperforming all baselines. 
As controls, gray, random noise, and white patches are also tested. 
Qualitative results on MS-COCO in Fig.~\ref{fig3:Attack Performance} further confirm the effectiveness of AdvAD.
We also evaluate transferability across datasets and detectors.
%Patches are trained on INRIA using YOLOv5s and Faster R-CNN, then tested on Pascal VOC, MS-COCO, and multiple victim detectors. 
% We assess transferability by training patches on INRIA with YOLOv5s and Faster R-CNN, then testing on Pascal VOC, MS-COCO, and various detectors.
% As reported in Table~\ref{t1}, AdvAD consistently yields low mAP across detectors, demonstrating strong transferability. 
AdvAD consistently attains low mAP across datasets and detectors, demonstrating strong transferability.
The quantitative and qualitative results show that our method achieves better attack performance.

\subsection{CARLA-Based Simulation Evaluation} 
\begin{figure}[ht!]
    \centering
    \includegraphics[scale=0.3]{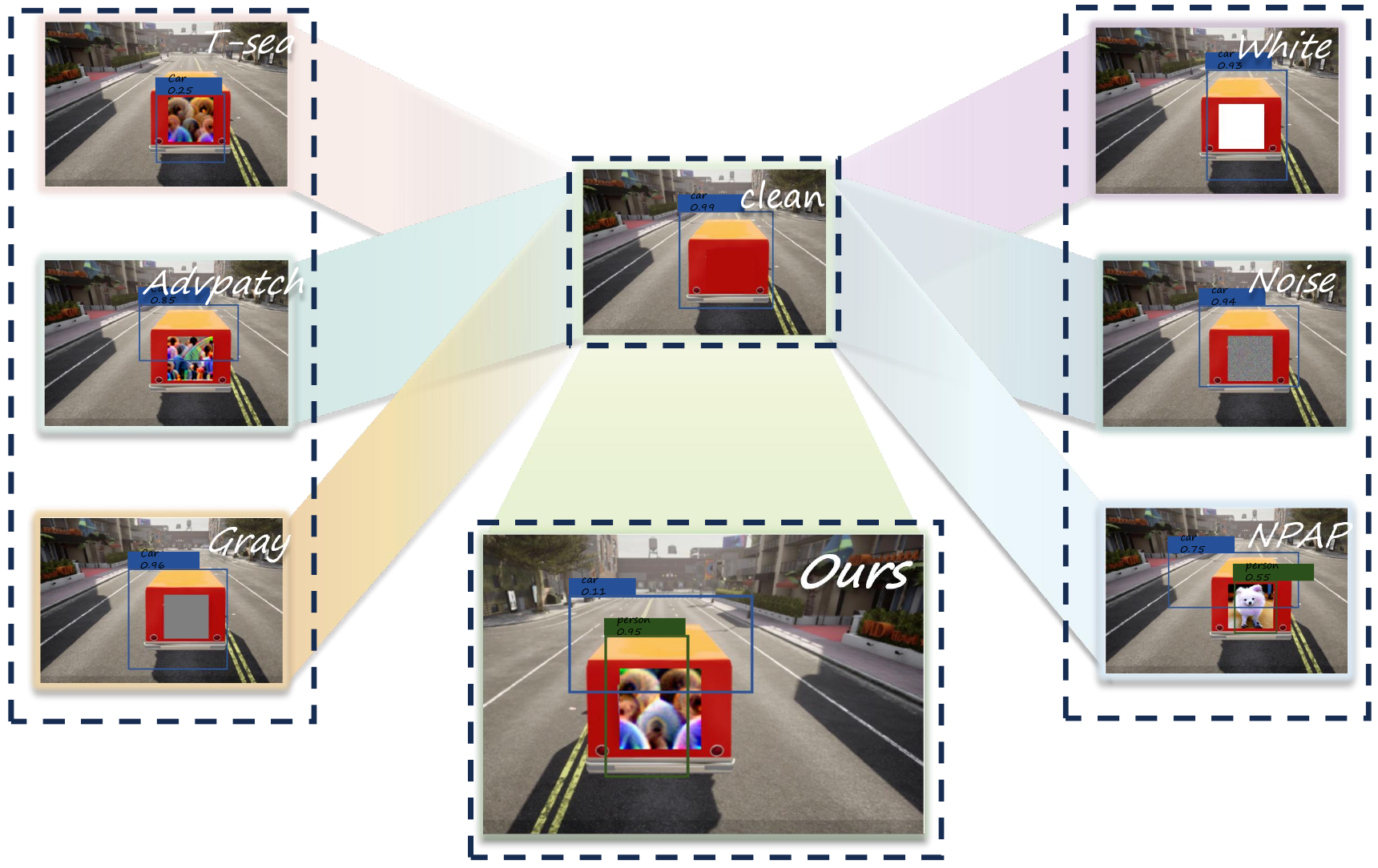}
    \caption{Visualization results in simulated environments. Notably, since the generated adversarial patches are regular in shape and suitable for planar placement, we select trucks as the target vehicles in our simulations.
    }
    \label{fig5:fangzhen demo}
\end{figure}

% \textbf{Results in simulated environments.} 
We evaluate AdvAD in autonomous-driving scenarios using the CARLA simulator~\cite{dosovitskiy2017carla} to emulate road elements and sensor settings. 
CARLA is an open-source simulator designed for the development, training, and validation of autonomous driving systems. 
It provides high-fidelity urban environments and supports diverse sensor simulations, making it widely used in research areas such as perception, planning, control, and adversarial robustness.
Scenes include stop signs, trees, pedestrians and vehicles; patches (rectangular and planar) are rendered onto the rear or side panels of trucks via planar projection. 
After deployment, we sample images from multiple viewpoints and feed them to the detector for evaluation. 
Qualitative examples appear in Fig.~\ref{fig5:fangzhen demo}, where AdvAD achieves stronger attacks: it not only mislocalizes bounding boxes but also induces false positives, leading to larger mAP degradation.

\subsection{Physical-World Evaluation} \label{sec:transferability}
\begin{figure*}[t!]
    \centering
    \includegraphics[scale=0.55]{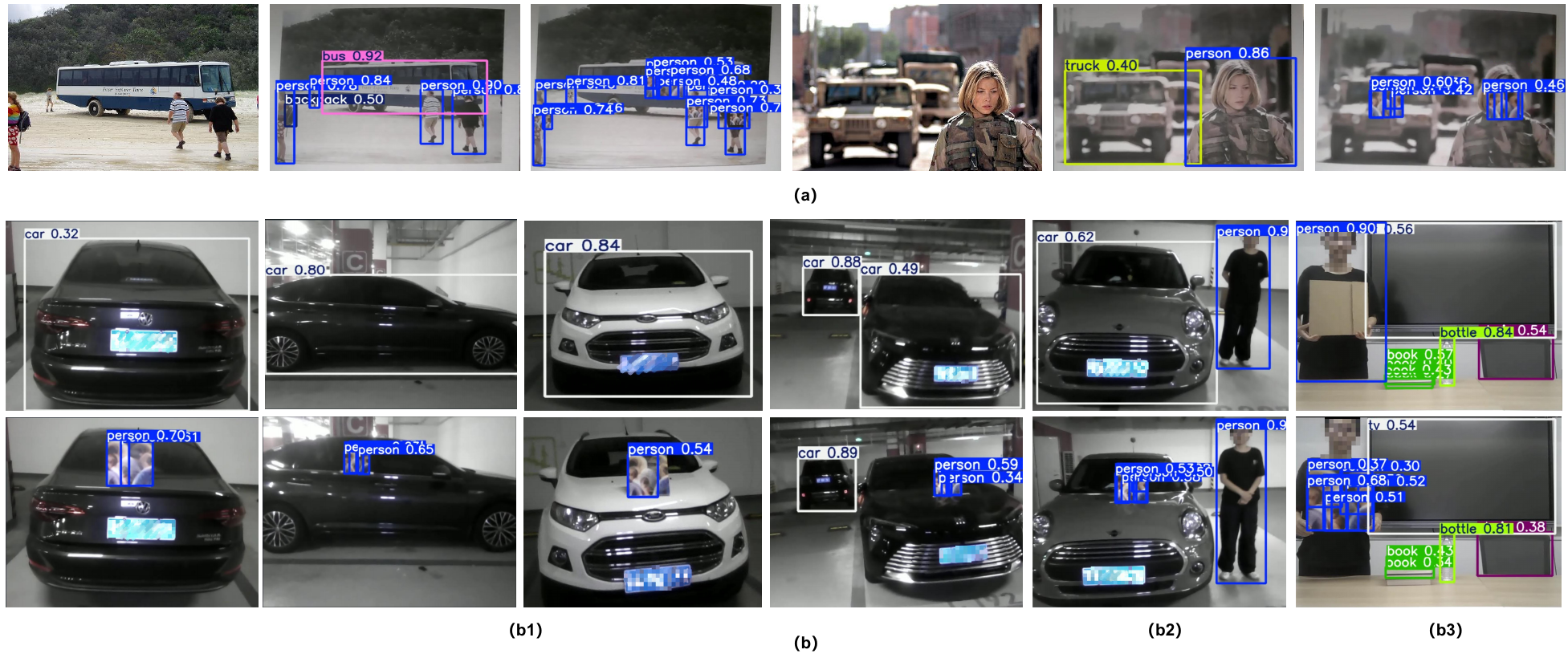}
    \caption{Visualization results in the physical world. (a) Physical-world validation under controlled conditions, where adversarial patches are printed and captured by a camera to evaluate attack effectiveness after physical rendering. (b) Real-scene deployment of printed adversarial patches under diverse environments. (b1) shows detection comparisons of vehicles before and after applying adversarial patches under outdoor scenes with various viewpoints; (b2) presents detection results in mixed human–vehicle scenes, comparing vehicles with and without adversarial patches in the same environment; (b3) illustrates indoor scenarios containing humans and other objects, where detection results of humans before and after applying adversarial patches are compared. These results demonstrate that AdvAD maintains robust attack performance under real-world variations in viewpoint, illumination, background, and scene composition.
    }
    \label{fig4:Physical}
\end{figure*}

We conduct two experiments to validate the effectiveness of AdvAD in the physical world. 
In Fig.~\ref{fig4:Physical}(a), the experiment evaluates the physical realizability of adversarial patches by printing adversarial examples with patches and capturing them using a camera.
This setting aims to verify whether the adversarial effect can be preserved after common physical transformations introduced by the printing and imaging process.
In Fig.~\ref{fig4:Physical}(b), the experiment evaluates the robustness of AdvAD in realistic deployment scenarios by deploying printed patches in real scenes and recording the results.
Overall, the physical-world evaluations demonstrate that AdvAD exhibits strong robustness and transferability when moving from digital simulations to real-world environments. Despite the presence of complex environmental variations, AdvAD consistently achieves high attack success rates, confirming its effectiveness in practical physical-world deployments.

\subsection{Transferability Discussion} \label{sec:discussion}
Transferability is a central objective of AdvAD.
In the digital domain (Sec.~\ref{sec:attack_performance}), adversarial patches jointly optimized on two source detectors achieve substantial performance degradation on unseen detectors and datasets, indicating effective cross-model and cross-dataset transferability.
Furthermore, the physical-world experiments (Sec.~\ref{sec:transferability}) show that the adversarial effectiveness of AdvAD is preserved after printing, imaging, and real-scene deployment.
These results indicate that AdvAD learns robust adversarial patterns that successfully transfer from digital simulations to physical-world scenarios, highlighting its strong digital-to-physical transferability.
This digital-to-physical transferability is particularly important for practical adversarial deployments.

\subsection{Comparison Study} \label{sec:compare}
We compare attack performance across ensemble configurations.
Table~\ref{t2} evaluates ensembles of three detectors: V5 (YOLOv5), SSD, and FR (Faster R-CNN). 
Ensembles tested include V5+SSD, V5+FR, SSD+FR, and V5+SSD+FR; all other settings are held constant for fair comparison.
In these experiments, we keep all other settings consistent to ensure a fair comparison.
The experimental results indicate that among the two-model ensemble approaches, V5+FR achieves the lower mAP value, suggesting a stronger attack effectiveness and transferability. 
Although the three-model ensemble achieves lower mAP, it incurs much higher optimization cost.
Balancing effectiveness and efficiency, AdvAD adopts the V5+FR ensemble.
\begin{table}[!t]   % twocolumn \begin{table*}
		\fontsize{8}{10}\selectfont    %{字体尺寸}{行距}
		\centering
		\caption{Attack results on attack performance across ensemble configurations.}
		\label{t2}
			\begin{tabular}{cccccc}
                \toprule[1pt]
                 \multicolumn{2}{c}{Methods}  
                & YOLOv5 & Faster R-CNN & SSD &Avg. \\
    \midrule[1pt]
     \multicolumn{2}{c}{ V5 + SSD }         & 33.09 & 33.87 & 21.00 & 29.32 \\
     \multicolumn{2}{c}{ FR + SSD }         & 28.15 & 28.30 & 29.03 & 28.49 \\
     \multicolumn{2}{c}{ V5 + FR + SSD }    & 22.91 & 29.40 & 30.65 & 27.65 \\
     % \multicolumn{2}{c}{ V5 + FR (Ours) }   & 24.49 & 29.97 & 33.38 \\
     \multicolumn{2}{c}{ V5 + FR (Ours) }   & 23.61 & 29.27 & 32.08 & 28.32 \\
    % \midrule[1pt]
    %  \multicolumn{2}{l}{ - Dynamic weight}          & 23.90 & 29.66 & 32.18 \\
    %  \multicolumn{2}{l}{ - Patch cutout }           & 23.84 & 29.90 & 33.10 \\
    %  \multicolumn{2}{l}{ - Classification loss }    & 23.32 & 34.23 & 30.29 \\
    %  \multicolumn{2}{l}{ - Total variation loss }   & 25.04 & 37.81 & 31.87 \\
    \bottomrule[1pt]
                \end{tabular}
		%}
	\end{table}   % twocolumn \end{table*}

\begin{table}[!t]   % twocolumn \begin{table*}
		\fontsize{8}{12}\selectfont    %{字体尺寸}{行距}
		\centering
		\caption{Ablation Study about modules of AdvAD.}
		\label{t3}
			\begin{tabular}{ccccc}
                \toprule[1pt]
                 \multicolumn{2}{c}{Methods}  
                & YOLOv5& Faster R-CNN& SSD \\
    \midrule[1pt]
     % \multicolumn{2}{l}{ - Dynamic weight}          & 0 & 0  \\
     % \multicolumn{2}{l}{ - Patch cutout }           & 0 & 0  \\
     % \multicolumn{2}{l}{ - Classification loss }    & 23.32 & 34.23  \\
     % \multicolumn{2}{l}{ - Total variation loss }   & 25.04 & 37.81  \\
     \multicolumn{2}{l}{ - Dynamic weight}          & 23.90 & 29.66 & 32.18 \\
     \multicolumn{2}{l}{ - Patch cutout }           & 23.84 & 29.90 & 33.10 \\
     \multicolumn{2}{l}{ - Classification loss }    & 23.32 & 34.23 & 30.29 \\
     \multicolumn{2}{l}{ - Total variation loss }   & 25.04 & 37.81 & 31.87 \\
    \bottomrule[1pt]
                \end{tabular}
		%}
	\end{table}   % twocolumn \end{table*}

\subsection{Ablation Study}
To investigate the contribution of each component in our proposed approach, we conduct an ablation study by selectively removing specific modules, as reported in Table \ref{t3}. 
% Removing the dynamic weight strategy leads to a noticeable performance drop across all detectors, indicating that adaptive weighting plays an important role in balancing different optimization objectives during training and improving attack stability on diverse models.
Removing the dynamic weighting strategy results in consistent performance degradation across all evaluated detectors, indicating that adaptive weighting is important for balancing multiple optimization objectives and stabilizing the training process across heterogeneous models. 
When the patch cutout module is omitted, the average attack performance decreases, indicating that patch cutout improves transferability by enhancing robustness to spatial variations.
% Omitting the patch cutout reduces transferability.
Furthermore, removing the classification loss weakens semantic alignment with the target class and degrades attack effectiveness.
Finally, omitting the TV loss produces noisy, less natural patches that hurt physical-world robustness.
Overall, the ablation results demonstrate that each component of AdvAD plays a complementary role in improving attack effectiveness, transferability, and physical-world robustness.
%When the dynamic weight mechanism is removed, the attack fails to adaptively adjust the patch's influence across different detectors, leading to a noticeable performance degradation. 
% Excluding the patch cutout operation results in reduced transferability. 
% Furthermore, removing the classification loss weakens the semantic alignment between the adversarial patch and the targeted class, making the attack less effective. 
%Finally, omitting the TV loss causes the generated patch to become noisy and less natural, which negatively impacts the physical-world effectiveness. 
% These results demonstrate that each component is necessary for enhancing the attack success rate and improving physical world mobility.

\section{Conclusions}
\label{sec:Conclusion}
In this paper, We present AdvAD, a transferable physical-world adversarial patch attack against object detectors in autonomous driving.
% In this paper, we propose AdvAD, a transferable physical-world adversarial patch attack against object detection models widely deployed in autonomous driving systems.
% AdvAD improves cross-model transferability by combining a multi-detector ensemble framework with a dynamic loss-weighting strategy that adaptively balances the influence of each model during optimization.
% To enhance robustness under real-world conditions, we incorporate a diverse set of data augmentations and geometric transformations that simulate practical environmental variations.
AdvAD enhances cross-model transferability through a multi-detector ensemble with dynamic loss weighting, and improves robustness via realistic deployment augmentation suite simulating real-world conditions.
% Extensive experiments in both digital and physical environments demonstrate that AdvAD consistently outperforms existing approaches in attack success rate and transferability across diverse detection architectures.
Experiments in digital and physical settings show that AdvAD consistently outperforms existing methods in attack success and transferability.
Our method provides a practical and effective framework for evaluating the vulnerability of perception modules in safety-critical scenarios.

% In this paper, we present AdvAD, a transfer-based physical-world adversarial patch attack targeting object detection models commonly used in autonomous driving. 
% By leveraging a multi-model ensemble strategy and a dynamic loss-weighting mechanism, AdvAD significantly improves the transferability of adversarial patches across different detection architectures. 
% To further improve robustness, we apply a diverse set of data augmentation and transformation strategies that simulate realistic environmental variations. 
% Extensive experiments conducted in both digital and physical settings demonstrate that AdvAD achieves state-of-the-art performance on multiple benchmarks, outperforming existing methods in terms of both attack success rate and cross-model generalization. 
% Our approach offers a practical and scalable solution for evaluating and strengthening the security of perception systems in safety-critical applications.

% \section*{Acknowledgements}
% Shengshan Hu's work is supported by the National Natural Science Foundation of China (Grant Nos. U20A20177, 62372196).
% Dezhong Yao's work is supported by the National Natural Science Foundation of China under Grant No. 62072204. 
% Shengshan Hu and Dezhong Yao are co-corresponding authors.

\bibliographystyle{IEEEtran}
\bibliography{refs}

\end{document}